\documentclass[10pt,twocolumn,letterpaper]{article}

\usepackage{cvpr}
\usepackage{times}
\usepackage{epsfig}
\usepackage{graphicx}
\usepackage{amsmath}
\usepackage{amssymb}

\usepackage[breaklinks=true,bookmarks=false]{hyperref}

\usepackage[utf8]{inputenc} 
\usepackage[T1]{fontenc}    
\usepackage{hyperref}       
\usepackage{url}            
\usepackage{booktabs}       
\usepackage{amsfonts}       
\usepackage{nicefrac}       
\usepackage{microtype}    
\usepackage{color}
\usepackage{epigraph}
\usepackage{multirow,bigdelim}
\usepackage{array}
\usepackage{caption}
\usepackage[leftcaption]{sidecap} 

\newcommand{\venue}[1]{{\scriptsize #1}}

\newcommand{\SK}[1]{{\color{blue}{#1}}}
\newcommand{\up}[1]{\vspace{#1}}

\newcommand{\best}[1]{{\bf\textcolor{red}{#1}}}
\newcommand{\sbest}[1]{{\bf\textcolor{blue}{#1}}}

\cvprfinalcopy 

\def\cvprPaperID{4635} 

\ifcvprfinal\fi
\begin{document}

\title{\vspace{-1em} Unsupervised Primitive Discovery for Improved 3D Generative Modeling}


\author{Salman H. Khan$^{*\ddagger}$\qquad Yulan Guo$^{\dagger\diamond}$\qquad Munawar Hayat$^{*\star}$\qquad Nick Barnes$^{\ddagger\triangleleft}$\\
$^{*}$Inception Institute of Artificial Intelligence, UAE; $^{\dagger}$National University of Defense Technology, China;\\ $^{\ddagger}$Australian National University, AU; 
 $^{\diamond}$Sun Yat-sen University, China; $^{\star}$University of Canberra, AU \\
 $^{\triangleleft}$Data61, Commonwealth Scientific and Industrial Research Organization, AU \\
{\tt\small salman.khan@inceptioniai.org}
}

\maketitle

\begin{abstract}
   3D shape generation is a  challenging problem due to the high-dimensional output space and complex part configurations of real-world objects. As a result, existing algorithms experience difficulties in accurate generative modeling of 3D shapes. Here, we propose a novel factorized generative model for 3D shape generation that sequentially transitions from coarse to fine scale shape generation. To this end, we introduce an unsupervised primitive discovery algorithm based on a higher-order conditional random field model. Using the primitive parts for shapes as attributes, a parameterized 3D representation is modeled in the first stage. This representation is further refined  in the next stage by adding fine scale details to shape. Our results demonstrate improved representation ability of the generative model and better quality samples of newly generated 3D shapes. Further, our primitive generation approach can accurately parse common objects into a simplified representation.
\end{abstract}

\section{Introduction}

\setlength{\epigraphwidth}{18em}
\epigraph{\it `The objects seen could be constructed out of parts with which we are familiar.'}{L.G. Roberts}

\vspace{-0.5em}
Computer vision in its early days saw the emergence of parts-based representations for object representation and scene understanding \cite{8573760}. As early as 1963, Roberts \cite{roberts1963machine} presented an approach to represent objects using a set of 3D polyhedral shapes. Subsequently, Guzman \cite{guzman1968decomposition} introduced a collection of parts that appear in generic line drawings and demonstrated how they can be used to recognize 2D curved shapes. The generalized cylinders based representation to describe curved objects of Binford \cite{binford1971visual} was a significant breakthrough. It was developed further, including a pioneering contribution by Biederman, who introduced a set of basic primitives (termed as \emph{`geons'} meaning geometrical ions) and linked it with the object recognition in human cognitive system \cite{biederman1985human}.

\begin{figure}
\centering
\includegraphics[trim= 0 0cm 4cm 0, clip, width=\columnwidth]{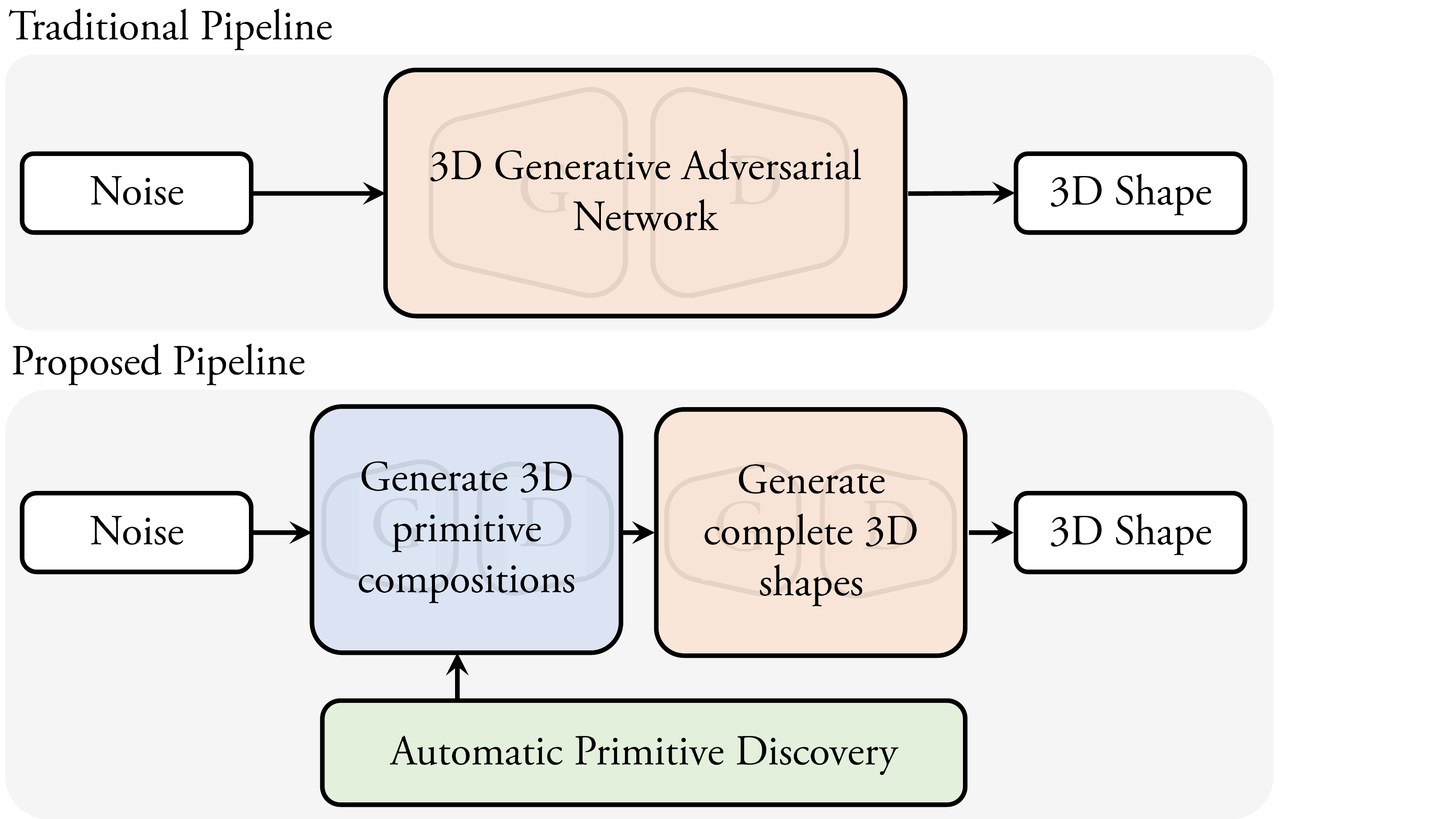}
\caption{Compared to traditional 3D generative modeling approaches (\emph{top}) that directly generate 3D shape, our approach (\emph{bottom}) transitions incrementally from a simple primitive based representation towards a complete 3D shape. Such a hierarchical approach provides better control and interpretability for generative networks. Furthermore, a major novelty of this work is an unsupervised primitive discovery approach that underpins the  proposed generative pipeline.}
\label{fig:overview}
\end{figure}

Very recently, early research towards automatic discovery of shape primitives using deep networks have been reported in the literature. Tulsiani \etal \cite{tulsiani2017abstraction} proposed a CNN model to predict the size and transformation parameters of primitives that were assembled together to represent generic 3D shapes. Their main draw-back is the inability to jointly represent different object categories using a single model. This requires a class-specific CNN training procedure, that is both time-consuming and difficult to scale to a large number of categories. Zou \etal \cite{zou20173d} proposed a generative model based on RNNs to recover a 3D shape defined by primitives from an input depth image. Their model, however, requires primitive-level shape labellings for training, requires an accurate depth map as input and works only for a set of three related classes (i.e., chair, table and night stand).

In this work, we propose to incorporate a generic primitive based representation in the 3D generative modeling process to enhance the scalability of learned models. Our first major contribution is the automatic primitive discovery in 3D shapes. Such a shape representation can provide several key benefits such as: \textbf{(a)} It factorizes the 3D generation process into a set of simpler steps, that defines a natural top-down flow in the existing bottom up generation pipelines. \textbf{(b)} It offers a highly compact representation compared to volumetric representations such as a voxel or a TSDF.  \textbf{(c)} Shape primitives provide a level of abstraction in the generation process, that makes it easy to understand and manipulate the output from generative models. \textbf{(d)} A global representation of a shape encoded by a few primitives allows a better intuition about the object parts, their physical properties (e.g., stability and solidness) and their mutual relationships (e.g., support and contact) \cite{hassanin2018visual}. \textbf{(e)} Such a shape description provides invariance to pose changes - by explicitly estimating object size and transformation using our proposed primitive Generative Adversarial Network (GAN), the network separates viewpoint changes from the actual shape changes.

In a nutshell, we introduce the principal of modularity in the existing generative pipelines. Our main contributions are summarized below:
\begin{itemize}\setlength{\itemsep}{0em}
\item A factorized generative model that improves 3D generation by introducing a simpler auxiliary task focused on learning primitive representation.
\item A fully unsupervised approach based on a high-order Conditional Random Field (CRF)  model to jointly optimize shape abstractions over closely related sub-sets of 3D models. Our model considers appearance, stability and physical properties of the primitives and their mutual relationships such as overlap and co-occurrence.   
\item The proposed model is jointly trained on all object categories and avoids expensive category specific training procedures adopted by earlier approaches. 
\end{itemize} 
 The proposed approach can be used to incorporate intermediate levels of user input and can render more sophisticated outputs on top of that. From another perspective, it can be used to analyze the intermediate part-based representations learned by GAN that provides better interpretability and transparent generation process. 


\section{Related Work}
\textbf{3D Generative Models:}
Wu \etal \cite{wu2016learning} were the first to extend the 2D GAN framework \cite{goodfellow2014generative} to generate 3D shapes. They demonstrated that the representations learned by the discriminator are generalizable and outperform other unsupervised classification methods. Another similar approach was proposed in \cite{smith2017improved} that used a Wasserstein loss \cite{arjovsky2017wasserstein}  for 3D GAN. However, \cite{smith2017improved,wu2016learning} do not address primitive based shape modeling for a hierarchical shape generation pipeline. Notably, some recent efforts in 2D image generation built a hierarchy of stacked GANs to generated stage-wise outputs \cite{huang2017sgan, wang2016generative,  zhang2017stackgan}. Huang \etal \cite{huang2017sgan} used a combination of encoder, generator and discriminator blocks to perform joint top-down and bottom-up information exchange for improved image generation. However, they operate on learned feature representations and do not enhance model interpretability. Besides, a common limitation of above mentioned methods is the lack of control over the latent representations and resulting difficulties in  generating data with desired attributes.

\textbf{Primitive Discovery:}
Cuboids have been extensively used in the previous literature to represent objects, parts and scene structural elements due to their simple form \cite{schwing2013box,khan2015separating,jiang2013linear,8573760}. The identification of recurring parts and objects has also been studied under the problems of co-segmentation and unsupervised learning \cite{rubinstein2013unsupervised,sidi2011unsupervised,rubio2012unsupervised}. In 3D shapes, some efforts aim at parts discovery and modeling their mutual arrangements in large-scale shapes datasets \cite{zheng2014recurring, fish2014meta}. Recently, Tulsiani \etal \cite{tulsiani2017abstraction} proposed a deep learning based approach to describe a shape with a combination of cuboid primitives.  Their approach requires learning a separate model for each set of  shapes belonging to the same category. Therefore, their model is not fully unsupervised and difficult to scale to a large number of object categories.  In this work, we address these limitations and further propose a factorized generative model for improved shape generation.

\textbf{Model Based 3D Reconstruction:} 
The pioneering work of Roberts \cite{roberts1963machine} lead to several efforts in recovering 3D layout of a scene from a single image. However, the 3D reconstruction from a single image is still an unsolved problem. Given the success of deep networks, recent approaches have proposed several incarnations of these models for 3D reconstruction. Izadinia \etal \cite{izadinia2016im2cad} generated 3D CAD models from a single indoor scene by detecting objects class and its pose using deep CNNs, followed by synthesizing scenes using CAD models from the ShapeNet library \cite{shapenet2015}. However, in contrast to these works, we do not have the prior knowledge about a specified set of primitives, rather we aim to automatically learn the shared parts across 3D shapes.

\begin{figure*}[htp]
\centering
\includegraphics[width=0.9\textwidth]{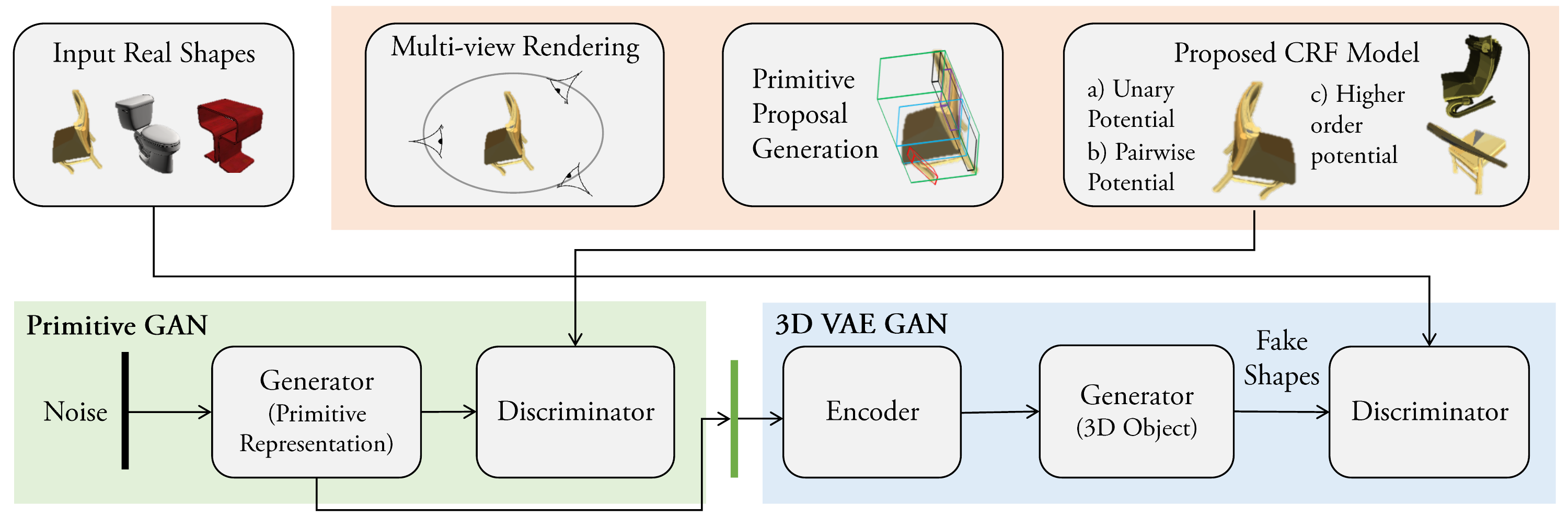}\up{-0.5em}
\caption{An overview of the proposed approach. Our model consists of a Primitive GAN that generates a parsimonious  representation that is used by the 3D VAE GAN in the next stage to recover a complete  3D shape.  }
\label{fig:overview}
\up{-0.5em}
\end{figure*}

\section{Primitive Discovery in 3D Shapes}
In the first stage, we automatically discover 3D primitives from generic object shapes. Our goal is to learn common recurring primitives in 3D shapes in an unsupervised manner. We introduce a higher-order CRF model that incorporates several physical and volumetric properties of primitives to identify a consistent shape description. We propose a multi-view primitive discovery approach that discretizes the 3D space without losing much shape information and allows a computationally efficient alternative to direct 3D primitive fitting. Furthermore, since our objective is to discover shared primitives among various models, direct cuboid fitting in the original 3D space leads to more instance specific and less category generalizable primitives. Our CRF model is explained next.

\subsection{Proposed CRF Model}
Our goal is to automatically discover 3D primitives to represent generic 3D shapes without any supervision. For this purpose, we design a CRF model that allows efficient inference and adequately incorporates rich relationships between primitives and complete shapes. Suppose, we have a dataset $\mathcal{D} = \{\mathbf{x}_1 \ldots \mathbf{x}_M \}$ consisting of $M$ 3D shapes. For each shape $\mathbf{x}_m$, assume a candidate set of box proposals generated via bottom-up grouping (see Sec.~\ref{sec:primgen}), denoted as $\mathcal{B} = \{\mathbf{b}_1 \ldots \mathbf{b}_N\}$, where $N$ is the total number of box proposals. The segmented regions obtained by grouping are denoted by $\mathcal{R} = \{\mathbf{r}_1 \ldots \mathbf{r}_R\}$. We also use a set of binary variables $\mathcal{V} = \{v_1 \ldots v_N\}$ and $\mathcal{S} = \{s_1 \ldots s_R\}$, where each $v_i$ and $s_r$ is associated with a box proposal and a segmented region, respectively. The variable $v_i$ denotes whether a cuboid is selected as a representative primitive or not.

We develop a CRF model to encapsulate the relationships between primitives both locally as well as globally.  The Gibbs energy formulation of the CRF is given by:
\begin{align}\small
E(\mathcal{V} | \mathcal{D}) = \sum_{i} \psi_u (v_i) + \sum_{i< j } \psi_p (v_i, v_j) 
+  \sum \psi_h (\mathcal{V}, \mathcal{T}), \notag
\end{align} 
where, $\psi_u, \psi_p $ and $\psi_h$ denote the unary, pairwise and higher-order potentials respectively and $\mathcal{T}$ represents the set of primitives from similar shapes. Next, we elaborate on each of the three potentials. 

\subsubsection{Unary Potential}
The unary potential for each primitive candidate denotes its likelihood for a valid simplified representation of the 3D shape. This potential encodes physical and geometric properties of each box. We explain the individual cost terms within the unary potential below.

\noindent
\textbf{Volumetric occupancy:} This cost ($c_i^{oc}$) estimates the empty volume within the $i^{th}$ primitive. It is defined as $c_i^{oc} = n^{oc}_i/n^{t}_i$, where $n^{oc}_i$ and $n^{t}_i$ are the number of empty and total voxels respectively. 

\noindent
\textbf{Shape uniformity:} This cost ($c_i^{su}$) measures the uniformity of the shape along the primitive sides that were used to propose the candidate primitive. It is calculated by taking the average entropy of the surface normal direction distribution for the relevant initial segmented regions. 

\begin{figure}[t]
\vspace{-1em}
    \centering
    \includegraphics[width=0.99\columnwidth]{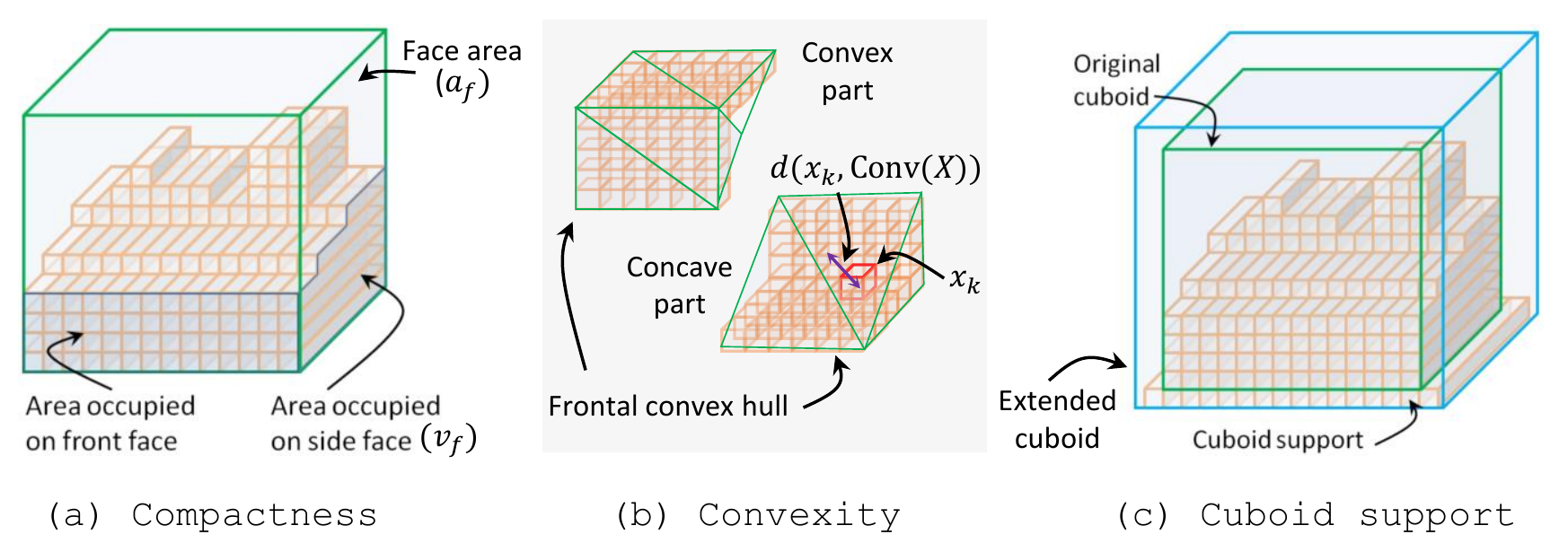}
    \caption{Visual illustration of costs.}
    \label{fig:rebut}
    \vspace{-1em}
\end{figure}

\noindent
\textbf{Primitive compactness:}  The cost ($c_i^{pc}$) estimates how tightly a 3D shape is enclosed by the primitive. It is calculated using average ratio between the empty surface area on each face and the actual face area ($a_f$): $\small c_i^{pc} = \sum_{f \in \mathcal{F}}\frac{a_f - v_f}{a_f},$ where $\mathcal{F}$ is the set of visible faces of primitive (Fig.~\ref{fig:rebut}).

\noindent
\textbf{Support cost:} A valid primitive is likely to be supported by nearby shape parts. This cost calculates nearby support by considering a $5\%$ enlarged box and taking the ratio: $c_i^{sc} = \frac{n^{sc}_i}{n^{ex}_i - n^{sc}_i}$, where $n^{ex}_i$ and $n^{sc}_i$ denote the number of voxels in the extended and original box primitive respectively.

\noindent
\textbf{Shape convexity:}
This cost determines the degree to which a shape-part is convex. For the regions associated with each primitive proposal, we first obtain a 3D frontal convex hull that only covers the visible 3D points from a single view. We then obtain the mean of distances between the 3D points and the frontal convex hull. It is given by:
     $c_i^{co} = \sum_{\forall\text{views}} \sum_k \frac{d(x_k,Conv(X))}{N},$
where, $x_k\in X$, $Conv(X)$ is the frontal convex hull for X, $d$ denotes shortest distance between $x_k$ and $Conv(X)$ (see Fig.~\ref{fig:rebut}). 
A large value of shape convexity cost ($c_i^{co}$) denotes that the shape is concave, while a small value denotes a convex shape. As convex shapes are more common in indoor scenes, a soft cost based on convexity is helpful.

\noindent
\textbf{Shape symmetry:}  For each primitive, we measure the cost ($c_i^{ss}$) denoting reflective symmetry of its enclosed 3D shape. For this purpose, we perform SVD decomposition to calculate three principal axis and measure the average overlap between the original points and their reflected versions. This overlap is measured as the distance between the neighboring point's position and normal direction \cite{karpathy2013object}. Given the three principal orthogonal directions $X = \{a,b,c\}$ of maximum variation and the corresponding Eigenvalues denoted by $\pi_x$, the following relation is used to measure symmetry:
\begin{align}\small
c_i^{ss} =  \frac{1}{\sum\limits_{x}\pi_x} \sum\limits_{x} \frac{\pi_x}{l_x} {\Big(} \sum_{j}{\parallel} p^{P^i_x}_j {-}
p^{P^i_{x'}}_{n_j}{\parallel} 
+ (1-q_j^{P^i_x} \cdot q_{n_j}^{P^i_{x'}}) \Big), \notag
\end{align}
where, $x{\in}X$, $j{\in}|P^i|$, $P^i$ denotes the point cloud of $i^{th}$ primitive, $P_{x'}$ denotes the flipped point cloud along $x$ direction, $p_j$ and $q_j$ denote the $j^{th}$ point and its normal respectively, $l_x$ denotes the length of primitive along the $x$ direction and $n^{j}$ is the nearest neighbor of $j^{th}$ point in $P_{x'}$. 

The individual costs listed above are fused together to obtain the per primitive unary cost as follows:
\begin{align}
\psi_u (v_i) &= \langle \mathbf{\mu}_{u}, \mathbf{w}\circ\mathbf{c}_i\rangle, \notag \\
\text{where }
\mathbf{c}_i &= [c_i^{oc}, c_i^{su}, c_i^{pc}, c_i^{sc}, c_i^{co}, c_i^{ss}].
\end{align}
Here $\langle\cdot,\cdot\rangle$, $\circ$ denote inner and Hadamard products, $\mu_u$ is the cost weight vector and $\mathbf{w}$ is the normalizing vector calculated on the validation set to obtain mutually comparable costs.



\subsubsection{Pairwise and High-order Potentials}
\textbf{Primitive Overlap:}
The pairwise potential considers the intersection relationships between primitive pairs. Since valid primitives do not significantly overlap each other, the goal is to penalize a configuration that violates this physical constraint. This cost $c^{pw}$ is measured as an intersection between the two cuboids normalized by the volume of the smaller cuboid:
$\psi_p(v_i , v_j) = \mu_{pw} c^{pw} v_i v_j, $
where $\mu_{pw}>0$ is the weighting parameter.
In practice, we introduce an auxiliary boolean variable $y_{ij}$ to linearize the pairwise intersection cost by replacing $v_i, v_j$ in the above cost. 

\noindent
\textbf{Primitive Parsimony:} 
Motivated by the minimum description length principle, we aim to obtain a parsimonious representation of 3D shapes. In other words, we discourage using additional primatives if a small number is adequate to represent an object.
A penalty on the number of active primitives is therefore introduced as a higher-order potential,
$\theta_h^{par} (\mathcal{V}) = \mu^{par} \sum_{i = 1}^{N} v_i, \; s.t., \; \mu^{par} > 0,$
where $\mu^{par}$ is the weight of the potential.

\noindent
\textbf{Coverage Potential:} 
The minimization of costs defined above will lead to a null primitive assignment. An important requisite is to obtain a representation that maximally covers the 3D shape \cite{tulsiani2017abstraction}.  This constraint is formulated as maximizing the surface area enclosed by primitives:
\begin{align}
& \theta_h^{cov} (\mathcal{V}, \mathcal{S}) = \mu^{cov} \sum_k c^{cov}_k s_k, \notag \\
&s.t., \; \mu^{cov} < 0, s_k \leq \sum_{i : \mathbf{r}_k \in \mathbf{b}_i} v_i.
\end{align}
Here, $\mu^{cov}_k$ denotes weight and the cost $c^{cov}_k$ is set equal to the area of the segmented region $\mathbf{r}_k$.

\noindent
\textbf{Co-occurrence Potential:}
We assume a set $\mathcal{T}$ of matched primitives for all vertices $v_i\in \mathcal{V}$. Each element $t_i  = \{\hat{v}_1 \ldots \hat{v}_J\} \in \mathcal{T} $ comprises of boolean variables $\hat{v}_j$ for all $J$ primitives identified in similar shapes that are matched to primitive `$i$'. The co-occurrence potential is defined as:
\begin{align}
& \theta_h^{coc}(\mathcal{V},\mathcal{T}) = \mu^{coc} \sum_{ij} c_{ij}^{coc} u_{ij}, \notag \\
& s.t., \; u_{ij} =  v_i  \hat{v}_j,\;  \mu^{coc} < 0,\; v_i \leq \sum\limits_{j} \hat{v}_j
\end{align}
The variables $u_{ij}$ and $\mu^{coc}$ denote the auxiliary boolean variable  and the weight respectively. The cost $c_{ij}^{coc}$ is defined as the Intersection over Union (IoU) measure between  $v_i$ and $\hat{v}_j$. We next describe the procedure used to find the set $\mathcal{T}$ for each 3D shape.

First, for each 3D volumetric object, a set of similar shapes is found via k-nearest neighbors in the feature space. The feature mapping is performed by obtaining a single 2D rendered image and feeding it forward through an off-the-shelf deep network \cite{simonyan2014very} pre-trained on the ImageNet dataset. Afterwards, we form a complete bipartite graph $\mathcal{G} = \{\mathcal{N}, \mathcal{E}\}$ with nodes $\mathcal{N}$ and edges $\mathcal{E}$. Assume that the capacity of each edge $e$ connecting nodes $p$ and $q$ is denoted by $w_{p,q} = -c_{e}$. The cost $c_{e}$ is defined by the IoU calculated for each edge in the bipartite graph. A canonical representation is obtained for 3D shapes by aligning their principal axes and matching spatial dimensions before primitive IoU calculation. The goal is to calculate maximum weight matching $\mathcal{M}$ between the disjoint partitions $\mathcal{P}$ and $\mathcal{Q}$ defined over the $m^{th}$ shape and its nearest neighbors. As a result, primitives within the 3D shape will have best matches that will preferably co-occur in similar 3D shapes. 
This problem can be formulated as an Integer Program (IP), but its solution is NP hard. To this end, we alternatively solve the following primal-dual linear relaxations of the original IP:
\begin{align}
\textbf{Primal:} & \;  \min \sum\limits_{p,q} w_{p,q}\, y_{p,q}, \quad
\text{s.t.}\; \sum\limits_{p \in \mathcal{P}} y_{p,q} = 1, \notag \\
&  \; \sum\limits_{q \in \mathcal{Q} } y_{p,q} = 1, \;  y_{p,q} \geq 0, \; p \in  \mathcal{P}, q \in \mathcal{Q}.
\end{align}
Since the relaxed LP does not guarantee an optimal solution, we also construct a dual to the original LP where both are solved alternatively to find the optimal matching. The following lower-bound is maximized in the dual formulation:
\begin{align}
\textbf{Dual:} \quad & \max  \sum\limits_{p\in \mathcal{P}}z_{p} + \sum\limits_{q \in \mathcal{Q}}z_{q} \notag \\
& \text{s.t.}\quad z_p + z_q \leq w_{p,q}, \quad (p,q) \in \mathcal{E}
\end{align}
The algorithm runs in several iterations maintaining a feasible solution to the dual problem, and tries to find a feasible solution to the primal problem that satisfies complementary slackness i.e., a perfect matching $\mathcal{M}$ with only tight edges \cite{schrijver2003combinatorial}.  Note that if the matching is not perfect, the exposed nodes in the graph do not have corresponding co-occurrence constraints during the final optimization.

\subsection{Model Inference}
For a given 3D shape dataset $\mathcal{D}$, the proposed CRF model represents each shape $\mathbf{x}_m$ with a set of primitive shapes. The CRF inference is formulated as a Mixed Integer Linear Program (MILP): 
\begin{align}
& \mathcal{V}^* = \underset{\mathcal{V}}{\text{argmin}} \;  E(\mathcal{V}|\mathcal{D}) \notag \\
\text{s.t.}\quad  & v_i = \{0,1\}, \; y_{i,j} \geq 0,  y_{ij} \leq v_i, y_{ij} \leq v_j, \notag \\
& y_{ij} \geq v_i + v_j - 1, \;\; s_k \leq \sum_{i : \mathbf{r}_k \in \mathbf{b}_i} v_i, s_k \leq 1 , \mu^{par}>0, \notag \\
& \mu^{par} > 0, \mu^{par} > 0, \mu^{cov} < 0,  \mu^{coc} < 0, \notag \\
& u_{ij} \geq 0, u_{ij} \leq v_i, u_{ij} \leq \hat{v}_j, u_{ij} \geq v_i + \hat{v}_j - 1, \;  \notag\\
& v_i \leq \sum_{j} \hat{v}_j \qquad   \forall i, \;\forall i,j, \; \forall k 
\end{align}
We use branch and bound algorithm \cite{land1960automatic} to efficiently solve the MILP based inference procedure.

\subsection{Primitive Proposal Generation}\label{sec:primgen}
Here, we describe our proposed multi-view approach to generate primitive candidates. Given a polygon mesh of a 3D CAD model, we obtain rendered depth views of the model from six equi-spaced virtual viewpoints around the object. The virtual camera viewpoints were divided into two groups, one looking horizontally at the center of the upright object and the second camera viewpoint was chosen at an upward elevation of $15^0$ such that the camera points towards the volume. These viewpoints were alternatively applied to get six rendered depth images that were subsequently used to obtain bottom-up polyhedron proposals. The rendered views provide sparse incomplete point clouds of the 3D shape that are mapped in the same frame of reference using a projective transformation.


As an initial step, we generate a set of 3D box proposals via bottom-up grouping. First, a normal image is calculated based on the 3D sparse point cloud for each view. Next, rough surface segmentations are obtained by clustering the 3D points that are co-located, have similar appearance and whose normals point in the same direction. 
Spurious segmented regions are removed by dropping regions with a small number of 3D points. We then calculate all closely lying region pairs, that can potentially form the two visible surfaces of a bounding box enclosing a part of the 3D shape. For each pair, a bounding box is tightly fit to generate a candidate primitive.

\section{Generative Modeling for Shape Generation}
The primitives discovered in an unsupervised manner allow us to factorize the shape generation process into two stages. The \textbf{first} GAN learns to generate novel primitive configurations that represent 3D shapes. The \textbf{second} GAN builds on this initial representation and fills in local details to generate a complete 3D shape. A Variational Auto-encoder (VAE) connects the two generative models. The overall pipeline therefore transitions from simple shape parametrization to more complex 3D shape generation. By introducing a simpler auxiliary task in the generative modeling, we achieve three key advantages: \textbf{(a)} In contrast to existing 3D generative models that are separately trained for each object category, our model is jointly trained on all shape classes, \textbf{(b)} It provides better interpretability of generator's latent space and can incorporate user input to generate desired shapes, \textbf{(c)} The learned model achieves better 3D generation results and demonstrates highly discriminative features. We explain the generative modeling pipeline below.

\begin{figure}[!htp]
\centering
\includegraphics[width=1\linewidth]{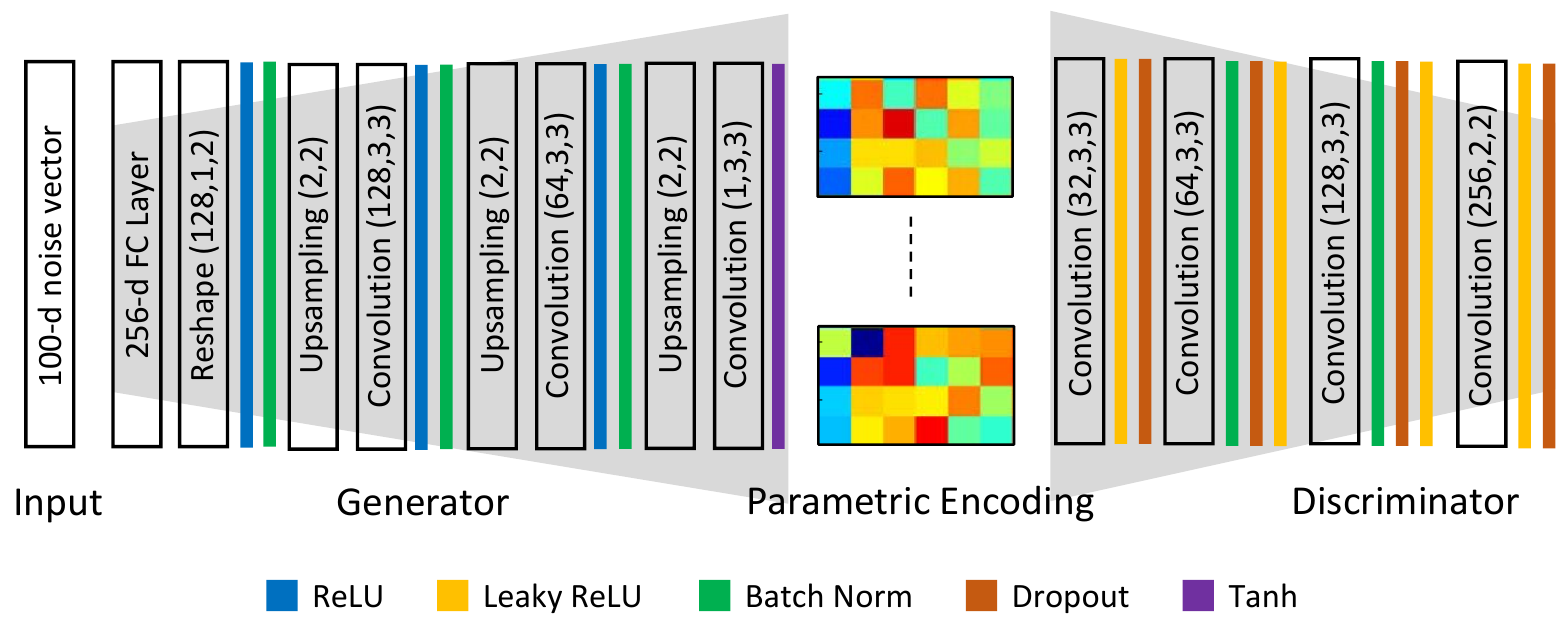}\up{-0.5em}
\caption{Primitive GAN Architecture.}
\label{fig:primGAN}
\up{-0.5em}
\end{figure}

\begin{SCfigure*}[][!htp]
\centering
\caption{3D Shape GAN Architecture. It first uses a VAE to encode the parametric representation of primitives and then learns to generate complete 3D shapes with an adversarial objective. Output tensors are shown with dotted lines.  }
\includegraphics[width=0.8\textwidth]{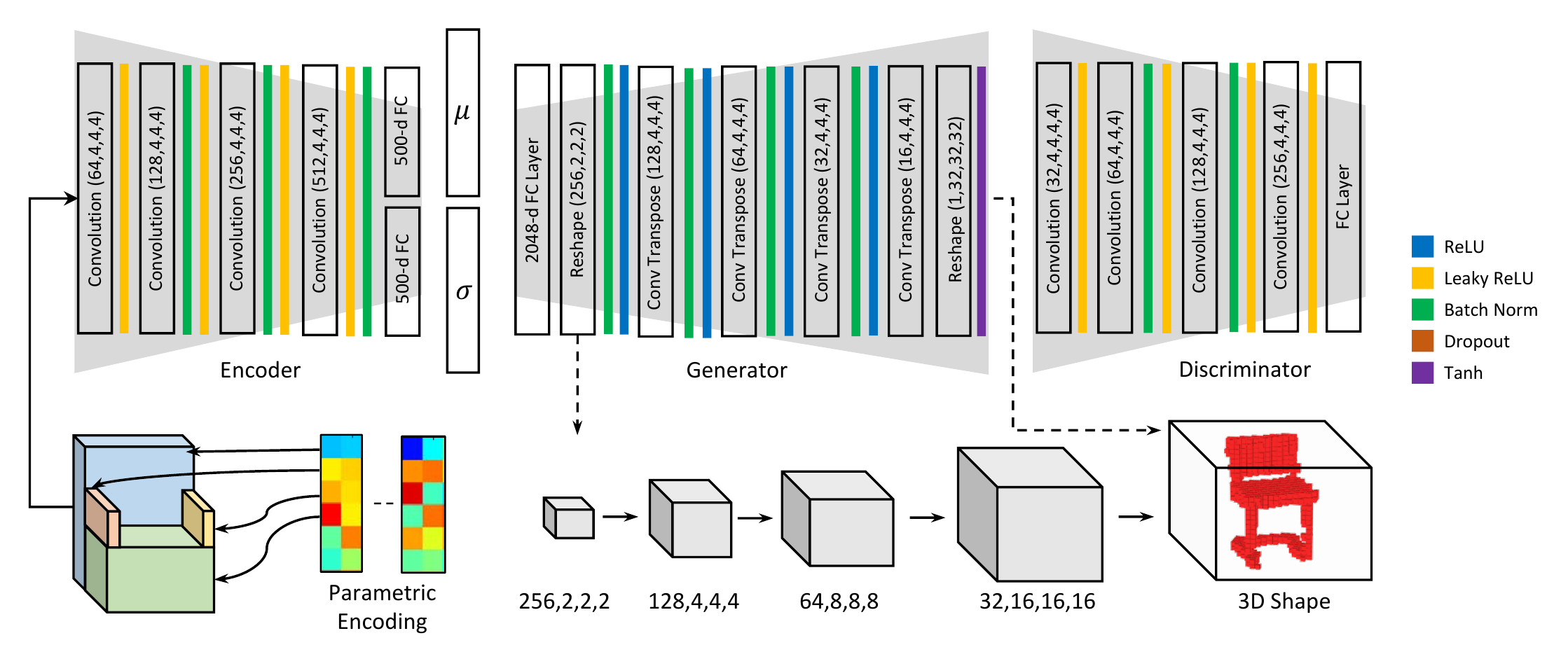}
\label{fig:shapeGAN}
\end{SCfigure*}

\subsection{Primitive GAN}

\textbf{Network Architecture:}
The primitive GAN consists of a generator and a discriminator network (Fig.~\ref{fig:primGAN}) \cite{khan2018guide}. The training process is framed as a game between the two competing networks. The generator maps a random sample (e.g., from a Gaussian distribution) to the original data space. The discriminator operates in the data space and predicts whether an input sample is real or fake.  The interesting aspect of our design is that an arbitrary number of primitives, $t \in [1,N]$ ($N=6$ in our case),  are predicted for each 3D shape. This flexibility is crucial because different object types are represented by different number of part primitives. $N$ can be
set higher at the cost of slower inference. Each primitive is encoded by a shape parameter set $\theta_s \in \mathbb{R}^{15}$ including the box dimensions (i.e., height, width, depth), translation (along x, y and z axes) and the rotation matrix (with nine parameters). An additional parameter $\theta_l$ is included to denote the likelihood whether the primitive will be selected in the overall shape or not. This likelihood is used as a parameter to obtain a sample from Bernoulli distribution that denotes the existence of a primitive \cite{tulsiani2017abstraction}. In this way, a 3D shape is encoded as a significantly lower dimensional parametric representation.

\textbf{Loss Function:}
To allow a stable training of GAN, we used an improved form of Wasserstein GAN \cite{arjovsky2017wasserstein}. The WGAN exhibits better convergence behavior by employing Wasserstein distance as an objective. It enforces the discriminator model to remain within the space of 1-Lipschitz functions by weight clipping that can lead to sub-optimal convergence behavior. Instead of weight clipping, we used the gradient penalty introduced in \cite{gulrajani2017improved} to restrict the norm of the gradients of the discriminator's output with respect to its input. The game between $D$ and $G$ is formulated as the following min-max objective function: 
\begin{align}\small
 \min_{G}\max_{D}  \mathop{\mathbb{E}}_{x\sim \mathbb{P}_r}[D(x)] - \mathop{\mathbb{E}}_{\tilde{x}\sim \mathbb{P}_g}[D(\tilde{x})] - \notag \\
 \lambda \mathop{\mathbb{E}}_{\hat{x}\sim \mathbb{P}_{\hat{x}}} [({\parallel}\nabla_{\hat{x}} D(\hat{x}){\parallel}_2 - 1)^2],
\end{align}
where $\mathbb{P}_r$ is the real data distribution, $\mathbb{P}_g$ is the generator distribution modeled as $\tilde{x}= G(z)$ such that $z$ is a random sample from a fixed distribution and $\mathbb{P}_{\hat{x}}$ is the distribution defined with uniformly sampled points between the pairs of samples belonging to $\mathbb{P}_r$ and $\mathbb{P}_g$.

\subsection{3D Shape VAE-GAN}
\textbf{Network Architecture:} The generative model for 3D shape generation consists of a combination of a variational auto-encoder and an adversarial network (Fig.~\ref{fig:shapeGAN}). The complete architecture consists of three blocks, an encoder, a generator and a discriminator. The parametric shape representation is first converted to a coarse 3D shape in the form of a voxelized grid. The encoder maps this representation to the parameters of a variational distribution  by applying a series of 3D convolutional and down-sampling operations.  The generator then operates on this a random sample from this parameterized distribution and generates a new 3D shape to deceive discriminator, while the discriminator is trained to correctly categorize the real and fake 3D shapes. Remarkably, in contrast to primitive GAN, the shape GAN consists of 3D operations to accurately model the data distribution of 3D shapes. 

\textbf{Loss Function:} The loss function has the same form as for the case of primitive GAN, however, a regularization is applied on the input latent representation of generator to match it to a fixed known distribution (a unit Gaussian).  This constraint is formulated as minimizing the Kullback-Leibler (KL) divergence between the Gaussian and encoded distribution as follows:
\begin{align}
\mathcal{L}_{vae} = \text{KL}(N(\mu, \sigma)||N(0,I)).
\end{align}
The reparametrization trick proposed in \cite{kingma2013auto} is used to perform back-propagation through the stochastic sampling from the distribution $N(\mu, \sigma)$.

\begin{SCfigure*}[][!htp]
\centering
{\includegraphics[width=0.617\textwidth]{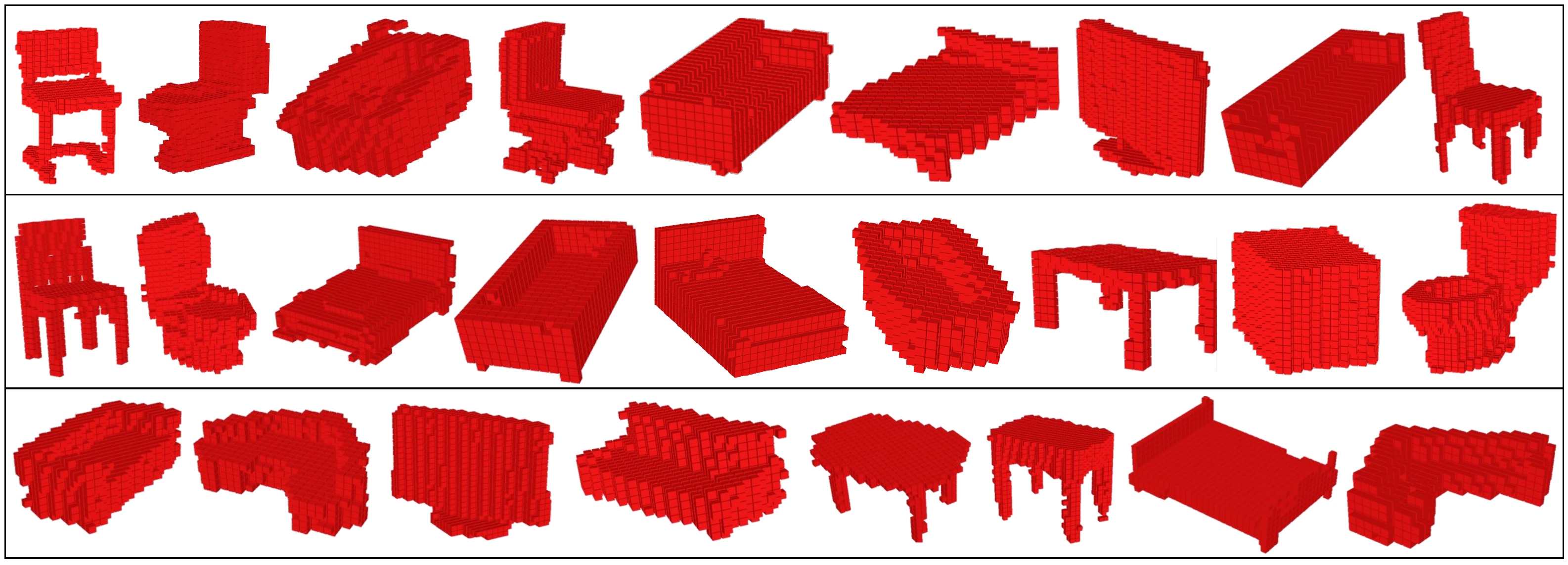}
\includegraphics[width=0.172\textwidth]{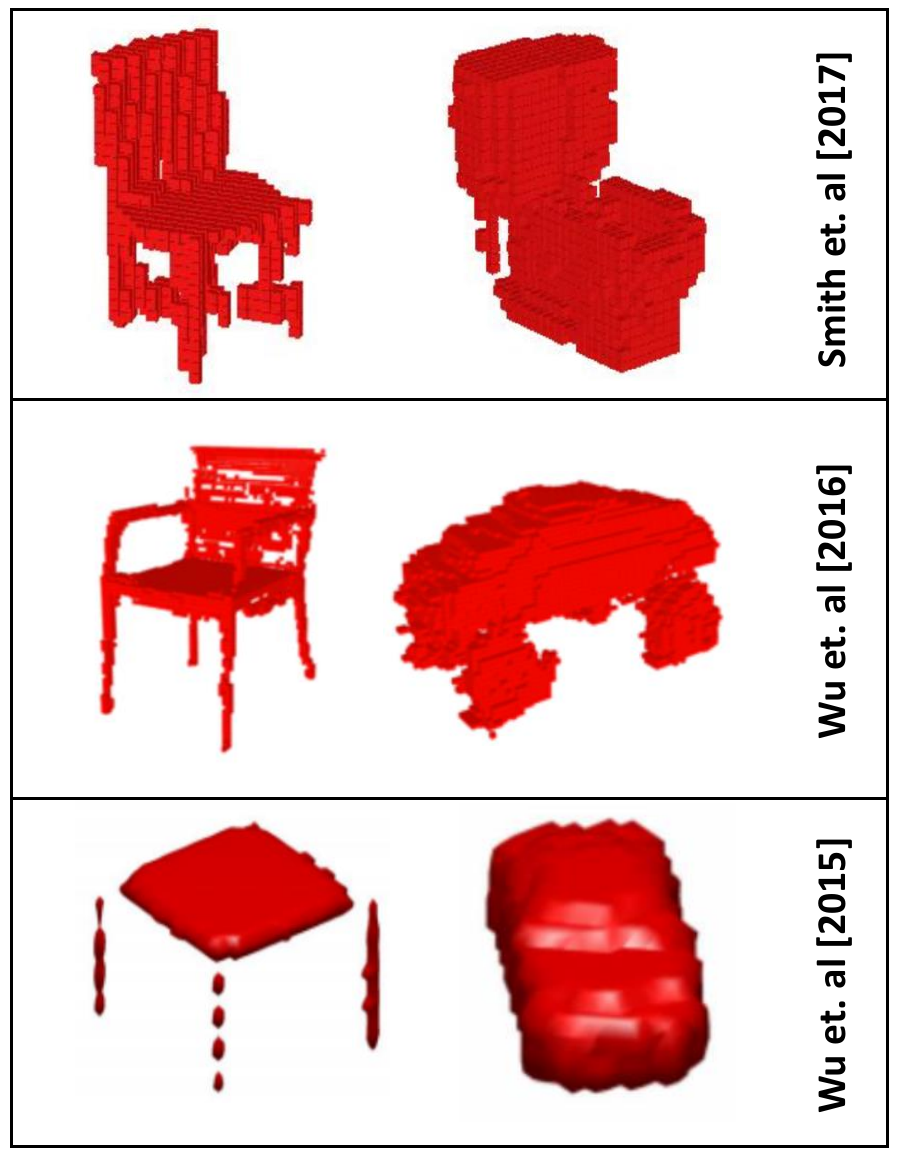}}
\caption{Qualitative results for 3D shape generation. \emph{Left}: Generated shapes from our model. \emph{Right:} Comparisons with  \cite{wu20153d}, \cite{wu2016learning} and \cite{smith2017improved} respectively from bottom to top.  }
\label{fig:qualfig}
\end{SCfigure*}


\begin{figure*}[!htp]
  \begin{minipage}[b]{0.34\textwidth}
    \centering
    \begin{tabular}{l c}\toprule
      Evaluation Measure & Performance \\ \midrule
        Recall & 83.0\% \\
        Precision & 19.9\% \\
        Accuracy & 60.8\% \\
        F-measure & 0.321 \\ \bottomrule
      \end{tabular}\vspace{-0.7em}
      \captionof{table}{Results for the primitive generation approach on ModelNet. A high recall shows that the predicted primitives generally tightly enclose the original 3D shape. }
      \label{tab:primacc}
    \end{minipage}
   \hfill
  \begin{minipage}[b]{0.65\textwidth}
  \centering
    \scalebox{1}{
\begin{tabular}{@{}l c  l c@{}}
\toprule
\multicolumn{2}{c}{\textbf{Supervised}} & \multicolumn{2}{c}{\textbf{Unsupervised}}  \\
 \midrule
 Methods & Accuracy & Methods & Accuracy \\
\midrule
PointNet \venue{(CVPR'17)} \cite{qi2017pointnet} & 86.2\% &  T-L Net \venue{(ECCV'16)} \cite{girdhar2016learning} & 74.4\% \\
OctNet \venue{(CVPR'17)} \cite{riegler2017octnet} & 83.8\% & 3D-GAN \venue{(NIPS'16)} \cite{wu2016learning} & 83.3\% \\
Vol-CNN \venue{(Arxiv'19)} \cite{ramasinghe2019volumetric} & 86.5\% & Vconv-DAE \venue{(ECCV'16)}\cite{sharma2016vconv} & 75.5\% \\
EC-CNNs \venue{(CVPR'17)}\cite{Simonovsky_2017_CVPR}   & 83.2\% &  3D-DescripNet \venue{(CVPR'18)} \cite{xie2018learning} & 83.8\% \\
\cmidrule{3-4}
Kd-Net \venue{(ICCV'17)}  \cite{Klokov_2017_ICCV} & 88.5\%  & 3D-GAN (Ours) & \sbest{84.5}\% \\
SO-Net \venue{(CVPR'18)} \cite{li2018so} & 90.8\% & Primitive GAN (Ours) & \best{86.4}\% \\
\bottomrule
\end{tabular}}\vspace{-0.7em}
\captionof{table}{Classification performance on the ModelNet40 dataset. }
\label{mn40_class}
  \end{minipage}
\end{figure*}

\begin{table}[!htp]
\centering
\scalebox{0.95}{
\begin{tabular}{c  c c }
\toprule
Type &  {Method} & Accuracy  \\
\midrule
 \multirow{5}{*}{\rotatebox{90}{Supervised}} & 3D ShapeNets \venue{(CVPR'15)} \cite{wu20153d} & 93.5\% \\
 & EC-CNNs \venue{(CVPR'17)} \cite{Simonovsky_2017_CVPR} & 90.0\% \\
 & Kd-Net \venue{(ICCV'17)} \cite{Klokov_2017_ICCV}  & 93.5\%  \\
 & LightNet \venue{(3DOR'17)} \cite{zhi2017lightnet} & 93.4\%  \\
 & SO-Net \venue{(CVPR'18)} \cite{li2018so} & 95.5\% \\
 \midrule
 \multirow{7}{*}{\rotatebox{90}{Unsupervised}} &  Light Field Descriptor \venue{(CGF'03)} \cite{chen2003visual} & 79.9\% \\
  & Vconv-DAE \venue{(ECCV'16)} \cite{sharma2016vconv} & 80.5\% \\
  & 3D-GAN \venue{(NIPS'16)} \cite{wu2016learning} & 91.0\% \\
  & 3D-DescripNet \venue{(CVPR'18)} \cite{xie2018learning} & \best{92.4\%} \\
  & 3D-WINN \venue{(AAAI'19)} \cite{huang20193d} & 91.9\% \\
  \cmidrule{2-3}
  & 3D-GAN (Ours) & 91.2\% \\
  & Primitive GAN (Ours) & \sbest{92.2}\% \\
\bottomrule
\end{tabular}}\vspace{-0.7em}
\caption{Classification performance on the ModelNet10. }
\label{mn10_class}
\end{table}

\section{Experiments}
\textbf{Primitive Discovery:} 
We evaluate the primitive detection accuracy on ModelNet10 dataset and report results in Table~\ref{tab:primacc}. Specifically, we convert the  shapes and the 3D primitive representations to a 50x50x50 voxelized output. Evaluation is performed by accounting for the matched voxel predictions for both outputs. We obtain a high recall rate of $83\%$ that confirms the correct enclosure of shape parts by primitives. In contrast, a lower precision is obtained because shape parts are often hollow, that give rise to unmatched empty voxels. In our case, recall is a more accurate measure to asses the quality of primitives. Example results for primitive generation are shown in Fig.~\ref{fig:qualfig}.

\textbf{Unsupervised Shape Classification:} To illustrate the improved performance of proposed generative model, we evaluate the representations learned by our discriminator (convergence plot is shown in Fig.~\ref{fig:loss}). A typical way of evaluating representations learned without supervision is to use them as features for classification. Note that the Primitive GAN model is only trained in an unsupervised manner on the ModelNet10 dataset, but tested on both ModelNet10 and ModelNet40 datasets. We extract intermediate feature layers from the discriminator, concatenate them and train a single layer neural network classifier. The classification results are shown in Tables~\ref{mn40_class} and \ref{mn10_class}. Our method beats all other unsupervised techniques by a fair margin of $1.9\%$ on ModelNet40 dataset. On ModelNet10, we achieve a competitive performance as compared to the state-of-the-art \cite{xie2018learning}. Note that some unsupervised methods have used class specific models, extra datasets (such as ShapeNet) and higher feature dimensions compared to ours. The proposed method also compares well with recent best performing fully supervised methods on both datasets. These approaches employ other tricks e.g., EC-CNNs \cite{simonovsky2017dynamic} performs voting over 12 views of each test model at test time.

\begin{SCtable}[0.6][!t]
    \centering
    \resizebox{0.65\columnwidth}{!}{
    \begin{tabular}{@{}c c c@{}}
    \hline
        Method & IS
    \\ \hline
    3D-ShapeNet \venue{(CVPR'15)} \cite{wu20153d} & 4.13$\pm$0.19\\
            3D-VAE  \venue{(ICLR'15)} \cite{kingma2013auto} & 11.02$\pm$0.42 \\
       3D-GAN \venue{(NIPS'16)} \cite{wu2016learning}  & 8.66$\pm$0.45\\
        3D-DescripNet \venue{(CVPR'18)} \cite{xie2018learning}  & \textbf{\color{red}11.77$\pm$0.42}\\
        3D-WINN \venue{(AAAI'19)} \cite{huang20193d} & 8.81$\pm$0.18\\
        Ours (Primitive GAN) &  \textbf{\color{blue}11.52$\pm$0.33} \\
    \hline
    \end{tabular}}
    \caption{Inception scores for 3D shape generation. {\color{red}Best} and {\color{blue}Second-best} scores are shown in color. }
    \label{tab:incep}
\end{SCtable}

\begin{SCfigure*}
\centering
\includegraphics[width=0.65\textwidth]{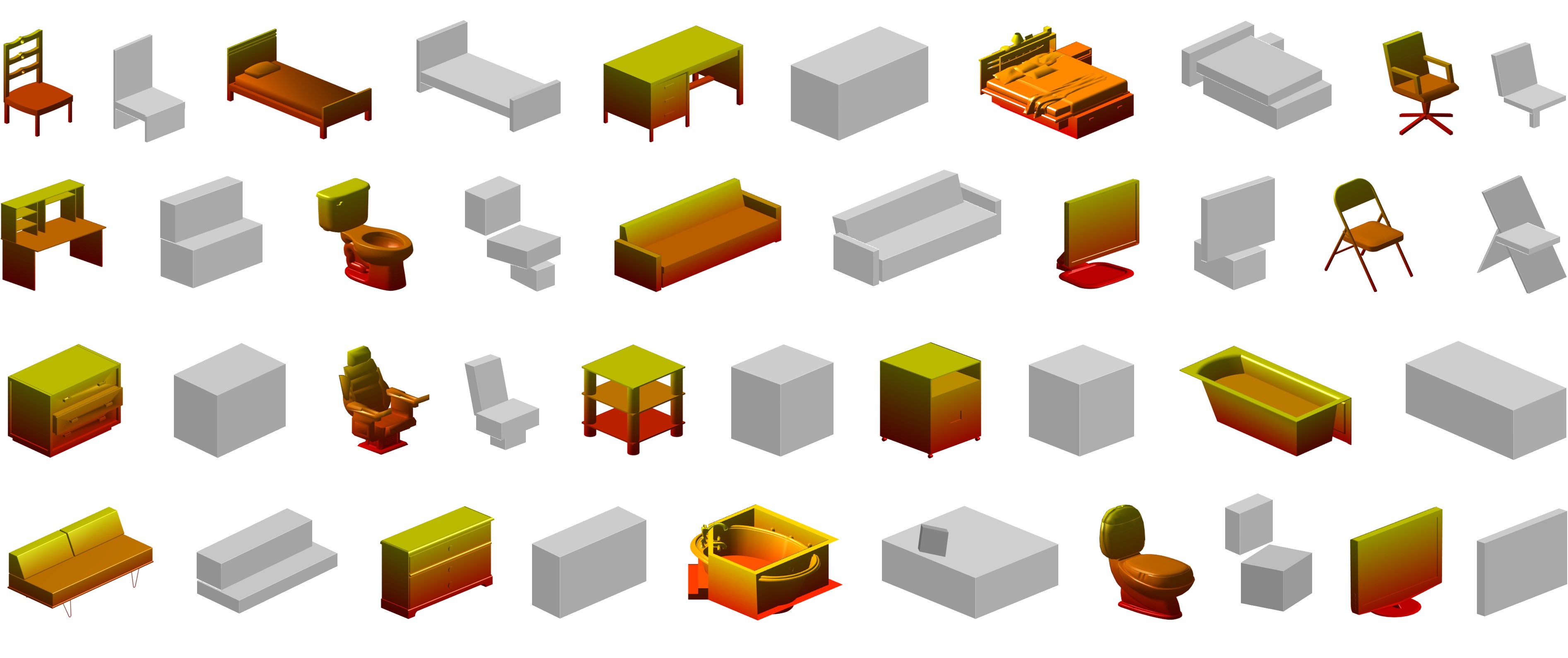}
\caption{Automatically discovered primitive representations of 3D shapes in an unsupervised manner. Example results are shown for common indoor objects such as chair, table, desk, bathtub, sofa, monitor, toilet and nightstand. Our approach learns to represent common shapes in a parsimonious form that is consistent for examples belonging to the same category. }
\label{fig:qualfig}\up{-0.5em}
\end{SCfigure*}

\begin{table*}[!htb]
    \begin{minipage}{.33\textwidth}
      \centering
    \includegraphics[trim= 0 0 4mm 8mm, clip, width=1\textwidth]{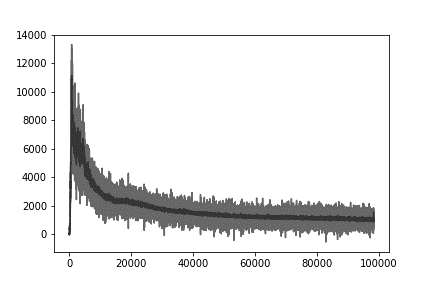}\up{-0.7em}
    \captionof{figure}{Discriminator loss during 3D GAN training on the ModelNet dataset. }
    \label{fig:loss}
    \end{minipage}%
    \hfill
    \begin{minipage}{.64\textwidth}
      \centering
      \scalebox{0.9}{
      \begin{tabular}{l  c c c c c c c}
\toprule
Method  &  Bed & Bookcase & Chair & Desk & Sofa & Table & Overall \\
\midrule
AlexNet-fc8$^{\dagger}$ \cite{girdhar2016learning} & 29.5 & 17.3 & 20.4 & 19.7 & 38.8 & 16.0 & 23.6\\
AlexNet-conv4$^{\dagger}$ \cite{girdhar2016learning} & 38.2 & 26.6 & 31.4 & 26.6 & 69.3 & 19.1 & 35.2\\
T-L Network$^{\dagger}$ \cite{girdhar2016learning} & 56.3 & 30.2 & 32.9 & 25.8 & 71.7 & 23.3 & 40.0\\
3D-VAE-GAN \cite{wu2016learning} &  49.1 & 31.9 & 42.6 & 34.8 & \textbf{79.8} & 33.1 & 45.2\\ 
VAE-IWGAN \cite{smith2017improved} & 65.7 & 44.2 & \textbf{49.3} & 50.6 & 68.0 & 52.2 & 55.0 \\
\midrule
Primitive GAN$^{*}$ &  \textbf{68.4} & \textbf{52.2} & 47.5 & \textbf{56.9} & 77.1 & \textbf{60.0} & \textbf{60.4} \\
\bottomrule
\end{tabular}}\up{-0.7em}
       \caption{Reconstruction results for voxel prediction on IKEA dataset (AP is reported). 
       $^{\dagger}$Accuracies are reported from \cite{wu2016learning}. $^{*}$Primitive GAN uses primitive representations obtained from 3D shapes and therefore has more supervision relative to compared methods that propose shapes from 2D images.}\label{tab:recons}
    \end{minipage} \vspace{-0.5em}
\end{table*}

\textbf{Inception Scores:}
To quantitatively evaluate the generated 3D shapes, we report Inception Score (IS) in Table~\ref{tab:incep}. The IS characterizes generated objects based on two distinct criterion: the quality of 3D outputs and their diversity \cite{salimans2016improved}.  The quality of generated outputs is measured by the conditional probability $p(y|\mathbf{x})$, where $y$ is the output label and $\mathbf{x}$ is the input shape. The diversity of a sample is computed by the marginal distribution $\int_{z} p(y|\mathbf{x}= G(z)) dz$. The KL-divergence between the two gives the Inception score: $\text{IS} = \exp(\mathbb{E}[\text{KL}(p(y|\mathbf{x}) || p(y))])$. Notably, using a single model for shape generation, our model achieves the second best IS score on ModelNet10 which denotes the diversity and the quality of generates shapes.


\textbf{Primitive Based Reconstruction:}
In order to evaluate the reconstruction performance of the proposed GAN model, we test our approach on the IKEA dataset (Table~\ref{tab:recons}). Previous works e.g., \cite{wu2016learning} aim to reconstruct a 3D model from a single color image. However, in our case, the VAE-GAN model is trained on parametric inputs representing a set of basic primitives instead of image inputs. Therefore, we first run our proposed primitive discovery algorithm on the IKEA dataset to estimate primitive representations and afterwards use these to reconstruct full shapes. Note that this dataset consists of 759 images with 1039 object crops and corresponding models that belong to six objects classes namely bed, bookcase, chair, desk, sofa, and table. Since the dataset shapes are at 20x20x20, we downscale the original network output to lower resolution for evaluation.

\textbf{Ablation study on cuboid detection:} It is important to note that the proposed CRF formulation is an integrated framework where several potentials are optimized jointly. For example, useful primitives that are shared across similar shapes cannot be detected if we exclude any of the co-occurrence, coverage, parsimony or overlap potentials. Intuitively, one can understand that potentials like shape coverage and parsimony have opposite goals and they balance each other to get an optimal representation.  However, we do run an ablation study with different types of unary potentials whose results are provided in Table~\ref{tab:ablation2} below. We also include a case where only unary costs are used to pick up the top four (average primitive number in dataset) primitives. We note that all potentials contribute to final performance and excluding one or some of them leads to lower recall rates. 

\begin{SCtable}[][!htp]
\centering
\scalebox{0.9}{
    \begin{tabular}{@{}l@{}  c@{}} \toprule
      Method & Recall \\ \midrule
      w/o Volumetric occupancy & 72.8 \\
      w/o Shape uniformity & 81.5 \\
      w/o Primitive compactness & 77.2 \\
      w/o Support cost & 82.9 \\
      w/o Shape symmetry & 80.0 \\
      \midrule
      Unary only (top 4 boxes) & 51.6\\
      Full model & 83.0 \\ \bottomrule
    \end{tabular}}\hspace{-0.5em}
   \caption{Ablation study on ModelNet10  for unsupervised primitive detection. The unary cost itself is insufficient to generate a good primitive representation. The best result is achieved with our full model.}
   \label{tab:ablation2}
\end{SCtable}


\section{Conclusion}
We factorized the generative image modeling problem to a set of simpler but interconnected tasks. Such a decomposition of problem allows GAN to generate realistic and high quality 3D voxelized representations. Our approach is motivated by the fact that common 3D objects can be represented in terms of a set of simple volumetric primitives, e.g., cuboids, spheres and cones. We first decompose a shape into a set of primitives that provide a parsimonious description with significantly less number of tunable parameters. Using this representation, we break-down the operation of GANs into simpler steps, that helps in learning better representations for data in an unsupervised fashion and makes it possible to easily incorporate user feedback if available. Such a high level supervision is helpful for complex image generation tasks such as 3D image generation and provides better interpretability and control over the outputs from GAN.


{\small
\bibliographystyle{ieee}
\bibliography{egbib}
}

\end{document}